\documentclass[preprint,12pt,authoryear]{elsarticle}

\usepackage{amsthm}
\usepackage{amsmath}
\usepackage{amsfonts}
\usepackage{bm,bbm}
\usepackage{algorithm,algpseudocode}
\usepackage{caption}
\usepackage{booktabs} 
\usepackage{graphicx}
\usepackage{enumerate}
\usepackage{amssymb}
\usepackage[colorlinks,citecolor=blue,urlcolor=blue,bookmarks=false,hypertexnames=true]{hyperref}

\usepackage{tikz}
\usepackage{amsmath}
\usepackage{amsfonts}
\usepackage{float}
\usepackage{booktabs}

\begin{document}

\begin{frontmatter}



\title{Disjoint principal component analysis by constrained binary particle swarm optimization}



\author[mymainaddress,mymainaddress3]{John Ram\'irez-Figueroa}
\ead{jramirez@espol.edu.ec}

\author[mymainaddress,mymainaddress3]{Carlos Mart\'in-Barreiro}
\ead{cmmartin@espol.edu.ec}

\author[mymainaddress,mysecondaryaddress]{Ana B. Nieto-Librero}
\ead{ananieto@usal.es}

\author[mymainaddress4]{Victor Leiva-S\'anchez}
\ead{victorleivasanchez@gmail.com}

\author[mymainaddress,mymainaddress3,mysecondaryaddress]{Purificaci\'on Galindo-Villard\'
	on}
\ead{pgalindo@usal.es}

\address[mymainaddress]{Department of Statistics, Universidad de Salamanca, Salamanca, Spain.}
\address[mymainaddress3]{ESPOL Polytechnic University, Escuela Superior Polit\'ecnica del Litoral, ESPOL, Facultad de Ciencias Naturales y Matem\'aticas, FCNM, Guayaquil, Ecuador.}
\address[mysecondaryaddress]{Institute of Biomedical Research of Salamanca, Salamanca, Spain.}

\address[mymainaddress4]{School of Industrial Engineering, Pontificia Universidad Cat\'olica de Valpara\'iso, Valpara\'iso, Chile.}

\begin{abstract}
In this paper, we propose an alternative method to the disjoint principal component analysis. The method consists of a principal component analysis with constraints, which allows us to determine disjoint components that are linear combinations of disjoint subsets of the original variables. The proposed method is named constrained binary optimization by particle swarm disjoint principal component analysis, since it is based on the particle swarm optimization. The method uses stochastic optimization to find solutions in cases of high computational complexity. The algorithm associated with the method starts generating randomly a particle population which iteratively evolves until attaining a global optimum which is function of the disjoint components. Numerical results are provided to confirm the quality of the solutions attained by the proposed method. Illustrative examples with real data are conducted to show the potential applications of the method.
\end{abstract}




\end{frontmatter}


\section{Introduction}

Most of multivariate techniques are based on the singular value decomposition (SVD) of the data matrix to be analyzed. This decomposition allows us to find a system of orthogonal principal or factorial axes, which correspond to the directions of maximum variance \cite{Beaton2014}. The vectors corresponding to these new axes are linear combinations of the original variables so that the new axes are latent variables. Each principal axis is related to a principal vector and the values of its coordinates or loads permit us to give practical sense to the axis according to the context of the research carried out \cite{HastieT2017}. This task can become a problem if a large number of coordinates have approximately equal absolute values, with no loads close to zero, making it difficult to characterize the axis corresponding to the latent variable to be considered. If each of the principal axes is a linear combination of a few original variables, their interpretation is simpler within the context of the investigation. Several multivariate techniques have been developed to achieve this goal. For example, the technique proposed in \cite{Vines2000}, which consists of obtaining directions that are represented by vectors of integers with several of their coordinates equal to zero; the sparse principal component analysis (PCA) proposed by \cite{Zou2004}, which seeks factorial axes with few non-zero loads; or the decomposition proposed by \cite{Mahoney2009a}, which consists of representing the data matrix as a product of three low-rank matrices. These techniques allow us to express the principal axes based on a reduced number of columns (variables) and/or rows (individuals or objects). Another proposal is to use disjoint components, which in addition to characterize a component as a linear combination of a few original variables, they do not appear in the other components, which justifies the name disjoint.

\cite{Vigneau2003} performed a hierarchical cluster analysis and then related a latent variable to each previously obtained cluster, proposing the disjoint PCA. In \cite{Vichi2009}, a different approach to the one used by \cite{Vigneau2003} was presented, determining the latent variables by means of the PCA, called clustering and disjoint PCA (CDPCA). The CDPCA consists of sequentially applying cluster analysis and PCA by means of an alternating least squares algorithm. \cite{Macedo2015} proposed an algorithm for the CDPCA. Another approach corresponds to the clustering and disjoint HJ biplot method proposed by \cite{Nieto-Librero2017}. Note that the disjoint PCA facilitates the interpretation of the results, without the need of existence of clusters to characterize them. \cite{NietoLibrero2015a} developed a disjoint HJ biplot method and its calculation algorithm is based on the approach proposed by \cite{Vichi2009} and the algorithm presented in \cite{Macedo2015}. \cite{Ferrara2016} introduced the three following methods to obtain disjoint components: (i) stepwise PCA; (ii) constrained PCA; and (iii) disjoint PCA, where the latter was derived from the CDPCA.

The present work is inspired by the disjoint HJ biplot method, introducing a different approach in the way how the optimization process is carried out. Here, a new paradigm in the optimization of data science problems is exhibited to minimize the objective function by means of an algorithm of the class of particle swarm optimization (PSO) of binary type with constraints. In addition, an alternative proposal is also presented to computate the eigenvalues associated with the inertia of each of the new principal axes, which enables us to construct indicators of the explicability percentages of each disjoint axis obtained. To understand the changes proposed in the way of approaching and solving the optimization problem with the proposed method, Section \ref{sec:2} describes how the CDPCA works. Optimization by PSO is an emerging technique of evolutionary computation developed by \cite{Kennedy1995}, based on stochastic optimization and inspired by the behavior of biological species, such as bees, birds and fish. PSO has been applied in many areas such as control systems, data mining, graphical computing, neural networks, optimization of functions in high-dimensional spaces, scheduling problems and robotics \cite{Eberhart2001, Alatas2008, Sha2008, Vasile2011}. In the context of PSO for multivariate analysis, \cite{Voss2005} presented a method to utilize PCA with PSO in which the system of principal axes is moved next to the swarm. To determine the directions of these axes in each iteration, PCA is used. \cite{Chu2011} introduced a method that employs PCA to handle the bounds of the feasible region. \cite{BiboM.2012} utilized PCA to reduce dimension and select the most important principal components to then apply PSO. Similarly, \cite{Zhao2014} derived an approach that uses PCA to determine efficient directions to be employed in PSO. \cite{Song2017} utilized PSO for global optimization in the presence of discontinuity. Other applications of PSO are in the cluster analysis, particularly in the detection of centroids and in the selection of the number of clusters \cite{VanDerMerwe2003, Esmin2012, Gajawada2012, Wang2018}. In the articles cited above, with the exception of the work on cluster analysis, multivariate techniques utilize PSO to improve its performance in prior or posterior stages. In the present article, a different proposal is made: the use of PSO is within the multivariate method, specifically in the solution of the optimization problem inherent to the calculation of the disjoint principal components.

This paper is organized as follows. Section \ref{sec:2} provides background about the DPCA method. In Section \ref{sec:3}, a description and implementation of the binary PSO algorithm with constraints is presented. Section \ref{sec:4} summarizes the new proposed method and exposes the advantages of using optimization by PSO  through simulations. In Section \ref{sec:6}, we illustrate our method with real data to show its potential applications. Finally, Section \ref{sec:7} discusses the conclusions of this study.

\section{The DPCA method}\label{sec:2}

\subsection{Background}

The $ I \times J$ matrix $\bm{X}$ in the DPCA is expressed as
\begin{equation} \label{2.1} 
\bm{X}= \bm{A}\bm{B}^{\top}, 
\end{equation} 
where $\bm{A}$ is the $I \times Q$ score matrix, which contains the coordinates of the individuals in the reduced $Q$-dimensional space of the disjoint components; $\bm{B}$ is the $J\times Q$ component loading matrix, which contains the coordinates of the variable $j$ in the reduced $Q$-dimensional space and is subject to the following constraints:
\begin{enumerate}
	\item[(i)] $\sum _{j=1}^{J}b_{jq}^{2} =1$, for $q=1,\ldots,Q$;
	\item[(ii)] $\sum _{j=1}^{J}\left(b_{jq}^{} b_{jr}^{} \right)^{2} =0$, for $q=1,\ldots,Q-1$ and $r=q+1,\ldots,Q$;
	\item[(iii)] $\sum _{q=1}^{Q}b_{jq}^{2} >0$, for $j=1,\ldots, J$.
\end{enumerate}
Constraint (i) indicates that each column has norm one. Constraint (ii) expresses that two different columns are orthogonal, while constraint (iii) establishes that no rows correspond to a vector of zeros. Since this is a low-range approximation, we have that $\bm{X}$ defined in \eqref{2.1} is now given by
\begin{equation} \label{2.2} 
\notag
\bm{X}= \bm{A}\bm{B}^{\top} +\bm{E}, 
\end{equation} 
where $\bm{E}$ is the error matrix, such that $\bm{E}= \bm{X}-\bm{A}\bm{B}^{\top}$. Minimizing the Frobenius norm squared of the error matrix $\bm E$ is equivalent to minimizing the objective function 
\begin{equation} \label{eq:fo} 
F(\bm{A}, \bm{B})=\| \bm{X}-\bm{A}\bm{B}^{\top}\| ^{2}= \| \bm{E}\| ^{2},
\end{equation}
which corresponds to the sum of residual squares. Then, we have the optimization problem given by
\begin{equation} \label{2.4} 
\left\{\begin{array}{cc} {\; \min {\kern 1pt} {\kern 1pt} F(\bm{A}, \bm{B})=\left\| \bm{X}-\bm{A}\bm{B}^{\top} \right\|^{2} \quad} & {} \\ {\quad \textrm{subject to}{\kern 1pt} \; \bm{A},\bm{B}\textrm{\; as\; described\; above.}} & {} \end{array}\right. 
\end{equation} 
Since $\|\bm{X}\| $ is constant, minimizing ${\| \bm{X}-\bm{A} \bm{B}^{\top}\| ^{2} }/{\| \bm{X}\| ^{2} } $ is equivalent to minimizing $\| \bm{X}-\bm{A} \bm{B}^{\top}\| ^{2} $, and therefore now $F(\bm{A}, \bm{B})={\| \bm{X}-\bm{A}\bm{B}^{\top}\| ^{2} }/{\| \bm{X}\| ^{2} } $, that is, the squared relative error as defined in \eqref{eq:fo} is used as the objective function. Note that $F$ measures the degree of fit of the low-range approximation. The method described below summarizes how $F$ is minimized to determine the number $Q$ of principal components, which generates a $Q$-dimensional subspace where the original data are projected so that each original variable contributes to only one component. Note that these disjoint components are orthogonal.

\subsection{Explanation of the method} 

The method used is an adaptation of that employed to construct the HJ biplot method \cite{NietoLibrero2015a} and the principal components \cite{Ferrara2016}, and  contains the following steps:
\newpage
\paragraph{Step $0$.} Let $\bm{X}$ be an $I\times J$ data matrix, with $Q$ being chosen as the number of disjoint components to consider. The $J\times Q$ stochastic binary matrix $\bm{{V}}_0$ is randomly generated by rows, such that its elements are given by
$$
v_{jq} =
\left \{
\begin{array}{ll} {1,} & {\textrm{if the variable}\, j\; \textrm{contributes to the component}\; q;} \\ 
{0,} & {\textrm{otherwise;}} \end{array}\right. 
$$
satisfying the constraints
\begin{equation} \label{ZEqnNum432494} 
\sum _{q=1}^{Q}v_{jq} =1,\; \; j=1,\ldots,J,
\end{equation} 
\begin{equation} \label{ZEqnNum540534} 
\sum _{j=1}^{J}v_{jq}^{} >0,\quad q=1,\ldots,Q,
\end{equation} 
where the expression defined in \eqref{ZEqnNum432494} ensures that there is a one and nothing more than one in each row, whereas the expression defined in \eqref{ZEqnNum540534} ensures that there are no columns filled with zeros. From \eqref{ZEqnNum432494} and \eqref{ZEqnNum540534}, we have that $\sum _{j=1}^{J}v_{jq}^{} v_{jr}^{} =0$, for $q\ne r$, which means the columns are orthogonal. 

\paragraph{Example 1.} Let $J = 5$ and $Q = 3$. Then,
$$
\bm{V}_{0} =\left(\begin{array}{ccc} {1} & {0} & {0} \\ {1} & {0} & {0} \\ {0} & {0} & {1} \\ {0} & {1} & {0} \\ {0} & {0} & {1} \end{array}\right).
$$ 
Note that $\bm{V}_0$ is a position matrix of the original data matrix, since it indicates how the original variables are distributed in the selected disjoint components. This must ensure that no component have all null loadings. In that case, the component with the largest number of assigned variables is randomly divided into two parts and then one of the halves is passed to the component that has the column of zeros.

Once $\bm{V}_0$ is calculated, the loading matrix $\bm{B}_{0} $ is obtained as follows. For each component $q=1,\ldots,Q$, a partition matrix $\bm{W}_{0q} $ is generated from the data matrix $\bm{X}$, with $\bm{W}_{0q} $ having $I$ rows, such as $\bm{X}$, and a number of columns corresponding to the variables indicated by the ones in the $q$-th column of $\bm{V}_0$, as indicated in Remark 1, where the matrix $\bm{W}$ is defined.

\paragraph{Remark 1} Note that the matrix $\bm{W}$ is generated from the original data matrix $\bm{X}$ taking those columns corresponding to variables for which a one is present in the matrix $\bm{V}$ at the associated component. For instance, if for the component 1, the non-null elements of each column in the matrix $\bm{V}_0$ are $1, j$ and $J$, then one must take the $1$-th, $j$-th and $J$-th columns. Such columns are those forming the matrix $\bm{W}_{01}$.  For this matrix $\bm{W}_{01}$, one applies the HJ biplot \cite{NietoLibrero2015a} calculating the coordinates for the variables. From these coordinates, one must choose the first column and such values are the coordinates of the variables $1, j$ and $J$ in the first component. The remaining variables have coordinates equal to zero for this component.

\vspace{-0.25cm}

\paragraph{Example 1} (continued) Recalling that $J = 5$ and $Q = 3$. Then,
$$
\bm{W}_{01} ={\mathop{\left(\begin{array}{cc} {x_{11} } & {x_{12} } \\ {x_{21} } & {x_{22} } \\ {\vdots } & {\vdots } \\ {x_{I1} } & {x_{I2} } \end{array}\right)}\limits^{X_{1} \quad \; X_{2} }},\quad \bm{W}_{02} ={\mathop{\left(\begin{array}{c} {x_{14} } \\ {x_{24} } \\ {\vdots } \\ {x_{I4} } \end{array}\right)}\limits^{X_{4} }},\quad \bm{W}_{03} ={\mathop{\left(\begin{array}{cc} {x_{13} } & {x_{15} } \\ {x_{23} } & {x_{25} } \\ {\vdots } & {\vdots } \\ {x_{I3} } & {x_{I5} } \end{array}\right)}\limits^{X_{3} \quad \; X_{5} }}.
$$ 

Note that each $\bm{W}_{0q} $ is expressed by its SVD in the form
\begin{equation} \label{ZEqnNum975304} 
\bm{W}_{0q} =\bm{R} {\bm{\Lambda} } \bm{T}^{\top}.
\end{equation} 
where $\bm{R}$ is a unitary matrix, $\bm{\Lambda}$ is a rectangular diagonal matrix with non-negative real numbers on the diagonal, and $\bm{T}$ is also a unitary matrix, assuming suitable dimensions for these matrices. The diagonal entries $\lambda_{i}$ of $\bm{\Lambda}$ are known as the singular values of $\bm{W}_{0q}
$. The columns of $\bm{R}$ and $\bm{T}$ are called the left and right singular vectors of $\bm{W}$, respectively. Thus, we obtain $Q$ SVDs of the form \eqref{ZEqnNum975304}. Once the matrix $\bm{W}_{0q}$ is obtained, the matrix $\bm{B}_{0} $ is constructed, with the $q$-th column  of $\bm{B}_{0} $ being the right singular vector corresponding to the highest singular value of the $q$-th SVD. This column contains the loadings corresponding to the variables considered in the $q$-th column of $\bm{V}_0$, with zeros in the positions that are not considered in that $q$-th column. Since $Q$ SVDs are considered, the number of columns of $\bm{B}_{0} $ is $Q$. Once the matrix $\bm{B}_{0}$ is computed, the coordinates for the objects are obtained as $\bm{A}_{0} = \bm{X}\bm{B}_{0}$, and then the objective function is calculated as $F_{0} (\bm{A}_{0},\bm{B}_{0} )$.

\vspace{-0.25cm}

\paragraph{Step $k$.} It is assumed that $k - 1$ steps of the algorithm have already been executed and then arrays $\bm{V}_{k-1}, \bm{A}_{k-1},\bm{B}_{k-1}$ of $F_{k-1} $ are available. We start by updating the matrix $\bm{V}_{k}$, letting the $j$-th row, with $j =1,\ldots,J$, locate a one in the $q$-th position of this row, for $q =1,\ldots, Q$, and zeros in the rest of the row. Maintaining unchanged the rest of the rows of $\bm{V}_{k-1} $, we obtain a new position matrix denoted by $\widehat{\bm{V}}_{kq} $. With this position matrix, we get the partition matrix $\widehat{\bm{W}}_{kq}$, the respective $\widehat{\bm{A}}_{kq}, \widehat{\bm{B}}_{kq} $ matrices and then the objective function $\widehat{F}_{kq} (\widehat{\bm{A}}_{kq},\widehat{\bm{B}}_{kq} )$ is evaluated. The position where $\widehat{F}_{kq}(\widehat{\bm{A}}_{kq},\widehat{\bm{B}}_{kq})$ is maximum assigns a one in that row. This process of relocating the ones in all the rows of $\bm{V}_{k-1} $ is continued and the resulting matrix is called $\bm{V}_{k} $. Since in each row the position containing the one runs through all the $Q$ entries of the $j$-th row, the number of SVDs to determine $\bm{V}_{k} $ is $J \times Q$. If during the process some column of $\bm{V}_{k} $ is generated with all its elements equal to zero, it acts in the same way as explained in the construction of $\bm{V}_{0} $. Once the position matrix $\bm{V}_{k} $ is computed, the corresponding matrices $\bm{W}_{kq}$ are obtained and from them the loading matrix $\bm{B}_{k} $ is also reached in the way already indicated. Hence, the matrix $\bm{A}_{k} $ of scores is evaluated at $\bm{A}_{k} =\bm{X} \bm{B}_{k}
$
and then $F(\bm{A}_{k},\bm{B}_{k}) = F_{k}$ is calculated to check whether the stopping criterion is attained as indicated below.

\vspace{-0.25cm}

\paragraph{Stopping criterion} Set a tolerance value $\varepsilon >0$ and if $\left|F_{k} -F_{k-1} \right|<\varepsilon $, then the algorithm is stopped. If this is the case, the algorithm is finished and the solution is attained in this step, which is considered as the final solution. If the stopping conditions are not reached, iterate until the stopping criterion is reached. Another stopping criterion is the total number of iterations to be performed.

The way in which $\bm{V}_{k} $ matrix is constructed is not flexible, since when descending by rows in the construction of this matrix, the positions that contain the one of each row are fixed and do not change once they are determined. Another option would be to consider all possible binary matrices for $\bm{V}_{k} $, but the number of possible partitions can be excessively high, even for small values of $J$ and $Q$, as shown in Table \ref{explosion} assuming Example 1.

\begin{table}[H] 
	\centering
	\small	\caption{Number of operations to be performed for the case of Example 1.}	\label{explosion}
	\begin{tabular}{lll} 
		\toprule 
		$J$& $Q=2$ & $Q=3$ \\ 
		\midrule 
		$10$ & $\left|S{\kern 1pt} \right|=1\, 022$ & $\left|S{\kern 1pt} \right|=55\; 980$ \\ 
		$15$ & $\left|S{\kern 1pt} \right|=32\, 766$ & $\left|S{\kern 1pt} \right|=14\, 250\, 606$ \\ 
		$20$ & $\left|S{\kern 1pt} \right|=1\, 048\, 574$ & $\left|S{\kern 1pt} \right|=3\, 483\, 638\, 676$ \\ 
		$30$ & $\left|S{\kern 1pt} \right|=1\, 073\, 741\, 822$ & $\left|S{\kern 1pt} \right|=2.058879109 \times 10^{14} $ \\ 
		\bottomrule 
	\end{tabular}	
\end{table}

\subsection{The algorithmic form of the method}

Algorithm \ref{alg:1} summarizes the DPCA method.

\begin{algorithm}[H]
	\caption{\small DPCA} \label{alg:1}
	\begin{algorithmic}[1]
		
		\State 
		Read $\bm{X}$, $Q$ and the number of iterations \texttt{nIter}.	
		
		\State
		Initialize with ${{\bm{V}}_{0}}$ and $k=1$.
		
		\State				
		For each iteration $k$ from 1 to \texttt{nIter}:
		
		3.1 Update ${{\bm{V}}_{k}}$.
		
		3.2 Obtain partition matrices ${{\bm{W}}_{kq}}$.
		
		3.3 Carry out an SVD of ${{\bm{W}}_{kq}}$.
		
		3.4 Construct ${{\bm{B}}_{k}}$ and ${{\bm{A}}_{k}}$.
		
		3.5 Calculate the objetive function ${{F}_{k}}=F({{\bm{A}}_{k}},{{\bm{B}}_{k}})$.
		
		3.6 Do $k:=k+1$.
		\State End and report the results for ${{\bm{A}}_{k}}$ and ${{\bm{B}}_{k}}$.
		
	\end{algorithmic}
\end{algorithm}

\section{The PSO method}\label{sec:3}

\subsection{Background}

The PSO method has the following individual and social behaviors:
\begin{enumerate}[(i)]
	\item Particles or individuals are attracted to food.
	
	\item At any moment, individuals know their closeness to food. The closeness is estimated through the so-called fit and is the value assigned by the objective function to the position where the particle is located.
	
	\item Each particle or individual remembers its closest position to the food. This is the individual's historical knowledge.
	
	\item Each particle shares its information about its closest position to food with the particles closest to it. This is the historical knowledge of its neighborhood.
\end{enumerate}
Through two rules of interaction, individuals adapt their behavior to that of the most successful individuals in their environment \cite{Imran2013}. The PSO method allows the particles to explore the set of feasible solutions $S$ in search of the optimum (food). The initial particle population is kept constant through the search process \cite{Eberhart2001}. At the instant of time $t$, each particle has a position ($\texttt{position}_{t}$), a velocity ($\texttt{velocity}_{t}$), and a value of fit $F_t$. The PSO algorithm iterates at each instant of time $t$ for each particle as follows:
\begin{enumerate}[(i)]
	\item \underline{Local information} ($\texttt{linformation}_{t}$): The current position with respect to which the particle reaches its best fit.
	\item \underline{Global information} ($\texttt{ginformation}_{t}$): The position where the neighborhood reaches the best fit.
\end{enumerate}
Each particle updates its position according to
\begin{equation} \label{ZEqnNum790844} 
\texttt{position}_{t} = \texttt{position}_{t+1}+\texttt{velocity}_{t+1},
\end{equation} 
where
\begin{equation} \label{ZEqnNum528449} 
\texttt{velocity}_{t+1} = \texttt{inertia}_{t}+\texttt{linformation}_{t}+\texttt{ginformation}_{t},
\end{equation} 
with $\texttt{inertia}_{t}$ being the component responsible for keeping the particle moving at the same direction in which it was moving until $t$; see more details about inertia in Remark 2.

\vspace{-0.25cm}

\paragraph{Remark 2} Note that, in the PSO context, inertia allows a better exploration of feasible solutions to be obtained. This inertia must not be confused with the inertia in the context of multivariate analysis, in which it is a measure of the geometric variability of the data set.

\vspace{0.25cm}

Recall that particles are represented by binary position matrices of the form $\bm{V}$ as defined in Section \ref{sec:2}. In each iteration, when updating the position, the particle leaves the set of feasible solutions $S$, since although the position matrix is binary, the velocity matrix is not. To solve the optimization problem given in \eqref{2.4}, the particles are represented by binary matrices $\bm{V}$ satisfying the constraints given in \eqref{ZEqnNum432494} and \eqref{ZEqnNum540534}. We must find the binary matrix $\bm{V}^{\star}$ that minimizes the objective function $F$ and therefore this is a binary optimization problem with constraints. The search for feasible solutions leads to a problem of high computational complexity (NP-hard) due to combinatorial explosion, as mentioned at the end of Section \ref{sec:2} and exemplied in Table \ref{explosion}. In the present article, we design of a constrained binary PSO algorithm to solve the combinatorial optimization problem associated.

\subsection{Explanation of the method}

The first step of the PSO algorithm is to initialize the particles, which means that they are placed at a random initial position. The number of particles $P$ is an important input parameter at this stage. The initial position of any particle is a feasible solution of the problem, that is, an element of a set of the feasible solutions $S$. In each iteration, the position and velocity of the particles must be updated according to the expressions given in \eqref{ZEqnNum790844} and \eqref{ZEqnNum528449}. The velocity matrices always have their components in the interval $[-1,\; 1]$, as shown in the continuation of Example 1 provided below. By updating the position at the stage $k + 1$, we have a new position that is not a binary matrix which satisfies the required constraints. 

\paragraph{Example 1} (continued)  Recalling that $J = 5$ and $Q = 3$. Let in the $k$-th step have position and velocity matrices given respectively by 
$$ \bm{V}_k =
\left(
\begin{array}{ccc} 
{1} & {0} & {0} \\ 
{1} & {0} & {0} \\ 
{0} & {0} & {1} \\ 
{0} & {1} & {0} \\ 
{0} & {0} & {1} 
\end{array}
\right),\quad 
\texttt{velocity}_k =
\left(
\begin{array}{rrr} 
{-0.51} & {0.75} & {0.12} \\ 
{-0.33} & {0.24} & {0.68} \\ 
{-0.83} & {-0.45} & {0.53} \\ 
{0.71} & {0.58} & {0.91} \\ 
{0.46} & {-0.79} & {0.64} 
\end{array}
\right).
$$ 
Then, when updating the position at the stage $k + 1$, we have a new position that is not a binary matrix which satisfies the required constraints given by
$$
\bm{V}_{k+1}  = \bm{V}_k + \texttt{velocity}_k = 
\left(
\begin{array}{rrr} {0.49} & {0.75} & {0.12} \\ {-0.33} & {0.24} & {0.68} \\ {-0.83} & {0.55} & {0.53} \\ {0.71} & {0.58} & {0.09} \\ {0.46} & {-0.79} & {0.64} \end{array}\right).
$$

Therefore, the proposed solution to the optimization problem given in \eqref{2.4} is as follows:
\begin{enumerate}[(i)]
	\item In each row, the entry with the highest absolute value is selected. The positions of these rows are occupied by ones and the rest by zeros. This procedure, as opposed to choosing random ones, allows the memory of the particle to be preserved.
	
	\item In the case where a column results only with zeros, choose the column with the highest number of ones, locate the one that came from the lowest absolute value, and change it to a column full of zeros. Thus, the matrix obtained satisfies the conditions of the set of feasible solutions $S$.
	
	\item If there are two columns full of zeros, apply the same procedure, that is, choose the two ones that come from the two smallest absolute values and then pass them to the columns of zeros.
	
	\item If there is a tie in the highest absolute values of a row, always choose the first of them from left to right.
\end{enumerate}
The points indicated in (i)-(iv) above as solution proposed, which convert a matrix with real inputs into a binary matrix in the set of feasible solutions $S$, are executed by a matrix operator denoted by $O$. The fit function is given by the squared relative error as defined in \eqref{eq:fo}, which represents, as in the DPCA method, the change for one unit. In the PSO algorithm there are two types of variables: individual and collective or global. In the individual ones, each particle stores its own values. In the global ones, the values are shared by all the particles. As PSO is an evolutionary algorithm, it may happen that, in two or more successive iterations, there is no change of the objective function. For this reason, the stopping criterion is the number of iterations given by the user \cite{Marti2009}.

In the sequel, $T$ is a matrix operator that transforms the binary matrix $\bm{V}$ into the loading matrix $\bm{B}$ as described in the step $k$ of Section \ref{2.2}. It is important to emphasize that, for calculating $\bm{B} = T(\bm{V})$, the number of SVDs performed by the operator $T$ is the same as the number $Q$ of desired disjoint components. Next, we describe how local and global informations work to reach the best fit.

\paragraph{\underline{Local information}}
As mentioned, it is the current position with respect to which the particle reaches its best fit. The local information of a particle quantifies its attraction to the position in which it obtains the best fit. For a specific particle $p$, the following information is stored in memory:
\begin{enumerate}[$\bullet$]
	\item
	$\bm{V}_{p} $ represents the binary matrix $\bm{V}$ for the particle $p$ in which it is located (current binary position).
	
	\item
	$\bm{B}_{p} =T(\bm{V}_{p})$ is the load matrix $\bm{B}$ for the particle $p$ in which it is located (current position).
	
	\item
	$F(\bm{V}_{p})$ corresponds to the value of the objective function evaluated at the current position of the particle $p$.
	
	\item
	$\bm{V}^{\star}_{p} $ is the best binary matrix $\bm{V}$ found by the particle $p$ locally.
	
	\item
	$\bm{B}^{\star}_{p} =T(\bm{V}^{\star}_{p})$ represents the best loading matrix found by the particle $p$ locally.
	
	\item
	$F(\bm{V}^{\star}_{p})$ is the value of the objective function evaluated at the best solution found by the particle $p$ locally.
	
	\item
	$\texttt{velocity}_{p} $ is the current velocity of the particle $p$, which is used to generate a disturbance of the current position of this particle to place it at a new position. Therefore, $\texttt{velocity}_{p} $ is a $J\times Q$ matrix, and as part of the algorithm, each entry in this matrix is in the interval $[-1, 1]$.
\end{enumerate}

\paragraph{\underline{Global information}}
As mentioned, it is the position where the neighborhood reaches the best fit. The global information quantifies the attraction of the particle towards its position at the best fit globally. In order to do this, the following information is collected at the level of the whole swarm:
\begin{enumerate}[$\bullet$]
	\item
	$\bm{V}^{\star}$ is the best binary matrix $\bm{V}$ found by the swarm of particles globally.
	
	\item
	$\bm{B}^{\star}=T(\bm{V}^{\star})$ is the best loading matrix found by the swarm of particles globally.
	
	\item
	$F(\bm{V}^{\star})$ represents the value of the objective function evaluated at the best solution found by the swarm of particles globally.
\end{enumerate}

As mentioned, the PSO algorithm has three main steps: initialization of the particles, iteration and update, which are detailed as follows:\\[0.25cm]
\textbf{Step 1 (initialization)}\\[-0.75cm]
\begin{enumerate}
	\item[1.1] For each particle $p$, from 1 to $P$:\\[-0.75cm]
	\begin{enumerate}[]
		\item 1.1.1 Generate the matrix $\bm{V}_{p} $ randomly.
		\item 1.1.2 Calculate $\bm{B}_{p} =T(\bm{V}_{p})$.
		\item 1.1.3 Fit $F(\bm{V}_{p})$.
		\item 1.1.4 Determine $\bm{V}^{\star}_{p} =\bm{V}_{p}$.
		\item 1.1.5 Compute $\bm{B}^{\star}_{p} =\bm{B}_{p}$.
		\item 1.1.6 Obtain the best fit $F(\bm{V}^{\star}_{p})$.
		\item 1.1.7 Establish $\texttt{velocity}_{p} $ randomly (initial velocity).
	\end{enumerate}
	\vspace{-0.5cm}
	\item[1.2] Among all the $P$ particles, search for the particle with the best binary initial position,
	which corresponds to the particle with the best fit, denoted by $p^{\star}$. Then:
	\begin{enumerate}[]
		\item 1.2.1 Assign $\bm{V}^{\star}=\bm{V}_{p^{\star}}$. 
		\item 1.2.2 Get $\bm{B}^{\star}={\bm{B}}_{p^{\star}} $ 
		\item 1.2.3 Obtain the best fit $F(\bm{V}_{p^{\star}})$.
	\end{enumerate}
\end{enumerate}
\vspace{0.25cm}
\textbf{Step 2 (iteration)}\\
The second step of the algorithm is the most important because it corresponds to the iterations. In each iteration all particles must move to a new position. Each position is a feasible solution to the problem, that is, it is not allowed a particle to place itself in the position of an unfeasible solution. In this stage, one has the following parameters:
\begin{itemize}
	\item[$\bullet$] \texttt{nIter}: number of iterations.
	
	\item \texttt{minIner}: it is the minimum inertia value, which corresponds to the value of the inertia in the last iteration.
	
	\item \texttt{maxIner}: it is the maximum inertia value, which is the value of the inertia in the first iteration.
	
	\item \texttt{wCognition}: it is the cognitive weight, which serves to control the cognitive component of the new velocity.
	
	\item \texttt{wSocial}: it is the social weight, which is used to control the social component of the new velocity.
\end{itemize}

\begin{enumerate}
	\item[2.1] To determine the new position of a particle $p$, it is necessary to calculate a new velocity that depends on three components: (i) the first of them is the velocity in the previous iteration that we control with inertia; (ii) the second one is the cognitive component that we handle with the cognitive weight; and (iii) the third  one is the social component that we control with the social weight. The inertia decreases linearly from iteration to iteration. The iterations are initiated with \texttt{maxIner}, whereas the iterations are finished with  \texttt{minIner}. Thus, for an iteration $0\le k\le \texttt{nIter}$, the inertia value is given by 
	\begin{equation} \label{4.3)} 
	\texttt{inertia} = \texttt{maxIner}-\left(\frac{\texttt{maxIner}-\texttt{minIner}}{\texttt{nIter}}\right)k. 
	\end{equation} 
	Expression \eqref{4.3)} implies, as it is iterated, that the new velocity depends less on the previous velocity, and the cognitive and social components affect it more. 
	
	\item[2.2] The value of the new velocity for a particle $p$ in the $k$-the iteration is obtained by 
	\begin{multline*}
	\small
	\texttt{newVel}_{p} = \texttt{inertia} \times \texttt{velocity}_{p} +\rho _{1} \times \texttt{wCognition} \times (\bm{B}^{\star}_{p} -\bm{B}_{p})\\+\rho _{2} \times \texttt{wSocial}(\bm{B}^{\star}-\bm{B}_{p})
	\end{multline*}
	where $\rho _{1},\rho _{2} \in [0, 1]$ are pseudo-random numbers so that $\texttt{newVel}_{p}$ has a stochastic component. Note that the matrix $\texttt{newVel}_{p} $ is of size $J \times Q$ and generally does not comply with the constraint that the entries are in the interval $[-1, 1]$. 
	
	\item[2.3] Therefore, a transformation must be applied to satisfy this constraint, for which we define the operator $L$ as follows.
	
	Let $z\in \mathbb{R}$ be any entry in the matrix $\texttt{newVel}_{p}$. The operator $L$ is defined by a convex linear combination of the sigmoid function, which is a particular case of the logistic function \cite{Nezamabadi2008,Sangwook2008} and given by
	$$
	L(z)=2\operatorname{sigmo}(z)-1,
	$$
	where $\operatorname{sigmo}(z)={{(1+e^{z})}^{-1}}\in [0, 1]$ so that $L(z)\in [-1,1]$. The operator $L$ is applied to all entries of $\texttt{newVel}_{p} $ as explained above, obtaining $\texttt{velocity}_{p}$ . 
	
	\item[2.4] Then, a new position (in floating point format) is calculated using
	\begin{equation} \label{4.7)} 
	\bm{V}^{(temp)}_{p} = \bm{B}_{p} + \texttt{velocity}_{p}.
	\end{equation} 
	Again, the operator $L$ is applied to $\bm{V}^{(temp)}_{p} $ in \eqref{4.7)}, so that its inputs are in the interval $\left[-1,1\right]$. Now, by applying the matrix operator $O$ to $\bm{V}^{(temp)}_{p}$, the new binary position of the particle $p$ is obtained in the set of feasible solutions as
	\begin{equation} \label{4.8)} 
	\notag	
	\bm{V}_{p} =O(\bm{V}^{(temp)}_{p}) 
	\end{equation} 
	\begin{equation} \label{4.9)} 
	\notag	
	\bm{B}_{p} =T(\bm{V}_{p}) 
	\end{equation}
	
\end{enumerate}

\textbf{Step 3 (updating)}
\begin{enumerate}
	\item[3.1] For each iteration $k$, from 1 to $\texttt{nIter}$:
	\begin{enumerate}
		\item[3.1.1] Calculate $\texttt{inertia}$.
		\item[3.1.2] For each particle $p$, from 1 to $P$ compute:
		\begin{enumerate}
			\item[3.1.2.1] $\texttt{newVel}_{p}$.
			\item[3.1.2.2] $\texttt{velocity}_{p} =L(\texttt{newVel}_{p})$.
			\item[3.1.2.3] $\bm{V}^{(temp)}_{p} =\bm{B}_{p} +\texttt{velocity}_{p}$.
			\item[3.1.2.4] $\bm{V}_{p} =O(\bm{V}^{(temp)}_{p})$.
			\item[3.1.2.5] $\bm{B}_{p} =T(\bm{V}_{p})$.
			\item[3.1.2.6] Fit $F(\bm{V}_{p})$.
			\item[3.1.2.7] If $F(\bm{V}_{p}) < F(\bm{V}^{\star}_{p})$, then
			\begin{enumerate}
				\item Determine $\bm{V}^{\star}_{p} =\bm{V}_{p}$
				\item Compute $\bm{B}^{\star}_{p} =\bm{B}_{p}$
				\item Compute $F(\bm{V}^{\star}_{p})$
			\end{enumerate}
			\item If $F(\bm{V}_{p}) < F(\bm{V}^{\star})$, then
			\begin{enumerate}
				\item Determine $\bm{V}^{\star}=\bm{V}_{p}$
				\item Compute $\bm{B}^{\star}=\bm{B}_{p}$
				\item  Compute $F(\bm{V}^{\star})$
			\end{enumerate}
		\end{enumerate}		
	\end{enumerate}
	\item[3.2]  Report the best solution found. 
\end{enumerate}

\subsection{Algorithm}

The construction of the CBPSO DC algorithm is summarized in Algorithm \ref{alg:2}. The $Step$ function places the particle $p$ in a new feasible position. The update of the position depends on the current position $Po{{s}_{p}}$ of the particle $p$, the cognitive weight $wC$, the best position found $BestPo{{s}_{p}}$ by the particle $p$, the social weight $wS$, the best position found by the set of particles $P$ and the current value of inertia. The $Step$ function is designed to control the feasibility of particle positions and allows exploring the search space intelligently.

\begin{algorithm}[H]
	\small
	\caption{\small
		CBPSO DC} \label{alg:2}
	\begin{algorithmic}[1]
		
		\State $p$ particles are randomly initialized		
		
		\State $\texttt{inertia} \leftarrow \texttt{maxIner},\text{ }BestPos\leftarrow Best(P)$
		
		\State For each iteration $k$ from 1 to $nIter$
		
		For each particle $p$ from 1 to $P$:
		
		$Po{{s}_{p}}\leftarrow Step\left( Po{{s}_{p}},wC,~BestPo{{s}_{p}},~wS,~BestPos,\texttt{inertia} \right)$
		
		If $f\left( Po{{s}_{p}} \right)<f\left( BestPo{{s}_{p}} \right)$ then $BestPo{{s}_{p}}\leftarrow Po{{s}_{p}}$ 
		
		If $f\left( Po{{s}_{p}} \right)<f\left( BestPos \right)$ then $BestPos\leftarrow Po{{s}_{p}}$ 		
		
		End For each $p$
		
		$\texttt{inertia} \leftarrow \texttt{maxIner}-\left(\frac{\texttt{maxIner}-\texttt{minIner}}{\texttt{nIter}}\right)k$
		
		End For each $k$
	\end{algorithmic}
\end{algorithm}

\subsection{Explained variance}

Note that the CBPSO DC method does not calculate the matrix $\bm{\Lambda}$. Then, the singular values of $\bm{X}$ are unknown and therefore the eigenvalues of \textbf{$\bm{X}^{\top} \bm{X}$}. In this work it is proposed to estimate the eigenvalues corresponding to each one of the disjoint components by means of the calculation of the variance that the individuals exhibit in the new referential system. This enables finding the percentage of explanation of the variance for each one of the axes of the disjoint referential system and the respective principal planes. In more detail, let $\bm{X}$ be a centered $I\times J$ data matrix, and let $\textrm{Var}(\bm{X})=\bm{S}$ be its variance-covariance matrix. Then the total variation of $\bm{X}$ is defined by

\begin{equation} \label{ZEqnNum916775} 
\textrm{tr}(\bm{S})=\sigma _{1}^{2} +\cdots +\sigma _{J}^{2} =\sum _{j=1}^{J}\sigma _{j}^{2}, 
\end{equation} 
where $\sigma _{1}^{2},\ldots,\sigma _{J}^{2} $ are the variances of the $J$ variables contained in the columns of $\bm{X}$, that is, $\textrm{Var}(X_{j})=\sigma _{j}^{2} $. Alternatively, we have that
\begin{equation} \label{ZEqnNum363554} 
\textrm{tr}\left(\bm{S}\right)=\alpha _{1} +\cdots +\alpha _{J}, 
\end{equation} 
where $\alpha _{1},\ldots,\alpha _{J} $ are the eigenvalues of $\bm{S}$.

Let $\bm{B}$ be a rotation matrix of size $J \times J$, that is, $\bm{B}$ is orthogonal in the sense that $\bm{B}^{\top} \bm{B}=\bm{I}_J$. The $\bm{B}$ column vectors define a new rotated referential system. Each new axis is a linear combination of the original variables and as such represents a new variable. Let $\bm{A}$ be a rotation transformation of $\bm{X}$ given by $\bm{A}=\bm{X}\bm{B}$. Its variance-covariance matrix is given by
\begin{equation} \label{4.12)}
\notag	 
\bm{\Sigma} =\textrm{Var}(\bm{A})=\bm{B}^{\top}\bm{S}\bm{B}.
\end{equation} 
Since the columns of $\bm{A}$ contain the coordinates of the individuals with respect to the new referential system, the variance of each column represents the variance of the new variables obtained by rotation $\bm{B}$. The total variation of $\bm{A}$ is defined as
\begin{equation} \label{ZEqnNum975538} 
\textrm{tr}\left(\bm{\Sigma} \right)=\textrm{tr}\left(\bm{B}^{\top} \bm{S}\bm{B}\right)=\textrm{tr}\left(\bm{S}\bm{B}\bm{B}^{\top} \right)=\textrm{tr}\left(\bm{S}\bm{I}\right)=\textrm{tr}\left(\bm{S}\right),
\end{equation} 
that is, the total variation of $\bm{X}$ and $\bm{A}$ is the same. Note that the rotation $\bm{B}$ does not affect the total variation. If $\beta _{1},\ldots,\beta _{J} $ are the eigenvalues of the matrix $\bm{\Sigma} $ in descending order, then the variance of the $j$-th new variable, denoted by $A_{j} $m is $\textrm{Var}(A_{j})=\beta _{j},$ for $j=1,\ldots,J$, and
\begin{equation} \label{ZEqnNum430899} 
\textrm{tr}\left(\bm{\Sigma} \right)=\beta _{1} +\cdots +\beta _{J}.
\end{equation} 
From \eqref{ZEqnNum916775}, \eqref{ZEqnNum363554}, \eqref{ZEqnNum975538}   and \eqref{ZEqnNum430899}, we have
\begin{equation} \label{4.15)} 
\notag		
\sigma _{1}^{2} +\cdots +\sigma _{J}^{2} =\alpha _{1} +\cdots +\alpha _{J} =\beta _{1} +\cdots +\beta _{J}.
\end{equation} 
Therefore, the coefficients $\beta _{i} $ can be used to determine the percentages of explanation of the disjoint axes. Thus, the percentage of the total variation explained by the $q$-th disjoint axis is given by
\begin{equation} \label{ZEqnNum234730} 
\frac{\beta _{q} }{\sum _{i=1}^{J}\beta _{i} } \times 100\% =\frac{\beta _{q} }{\sum _{j=1}^{J}\sigma _{j}^{2} } \times 100\% 
\end{equation} 
The $\sum _{j=1}^{J}\sigma _{j}^{2} $ amount can be obtained directly from the matrix $\bm{X}$ and the coefficients $\beta _{j} $ can be estimated from $\textrm{Var}(A_{j})$. In the case $Q=J$, when applying the CBPSO DC algorithm to obtain the disjoint components, the loading matrix $\bm{B}=\bm{I}_{J} $ is obtained and therefore $\bm{A}=\bm{X}\bm{B}=\bm{X}$. If $Q<J$, the columns of $\bm{B}$ are disjoint and of norm one. Therefore, they form an orthonormal set of $Q$ vectors in $\mathbb{R}^{J} $, and in addition $\bm{B}^{\top} \bm{B}=\bm{I}_{Q} $. Hence, $\bm{B}$ can be considered as a projection matrix in the new rotated disjoint referential system and $\bm{A}=\bm{X}\bm{B}$ is the matrix containing the coordinates of the $I$ objects found in the vector subspace generated by the first $Q$ disjoint components. As mentioned earlier, our proposal is to use the variance of scores, that is, we can apply \eqref{ZEqnNum234730}. The variance of the columns of $\bm{A}$ is used to calculate the variance explained by each disjointed axis obtained, and for each principal plane required. This criterion for calculating the variance of the columns of the scores matrix is used to determine the proportion of variability explained by each factor axis, both in the DPCA method and in CBPSO DC method in the numerical examples in the following sections. Note that is not required to calculate the matrix of singular values $\bm{\Lambda}$.

\section{Computational and simulation aspects}\label{sec:4}

In this section, we summarize the method in an algorithm and show the benefits of the CBPSO DC algorithm, illustrating with two kinds of examples the quality of the solutions that can be found. The first class considers two simulated examples and the second class presents two real data examples. The stop criterion used in the CBPSO DC algorithm is the number of iterations. However, each time the algorithm finds a better solution, it is stored along with its processing time. The best solution found by the algorithm usually has a processing time less than the time needed for all iterations to run.

\subsection{Summary of the method}

Algorithm \ref{alg:3} presents in a synthesized form the proposal given in Section \ref{sec:3} for calculating the disjoint components according to the CBPSO DC method.

All computational experiments were performed on a 64-bit Windows 10 computer, 8 GB of RAM, and an Intel (R) Core (TM) i7-4510U 2-2.60 GHz processor. The algorithm was implemented employing \texttt{C\#.NET} and \texttt{R}. The use of the \texttt{R} statistical software is important primarily for performing the SVD. More specifically, the \texttt{irlba} package was used as it provides a fast and efficient way to calculate a partial SVD for large matrices, instead of using the \texttt{svd} function of the \texttt{svd} package of \texttt{R}, which provides a generic SVD of a matrix. Communication between \texttt{C\#.NET} and \texttt{R} was possible using the \texttt{R.NET} middleware, which was installed in the corresponding code project as a \texttt{NuGet} package. The data matrix is located in an \texttt{Excel} sheet and is read from \texttt{C\#.NET} code at runtime using \texttt{COM+}. 

\begin{algorithm}[H]
	\small
	\caption{\small
		CBPSO DC} \label{alg:3}
	\begin{algorithmic}[1]
		
		\State Read a centered data matrix $\bm{X}$ of size $I\times J$, where $I$ is the number of individuals and $J$ is the number of variables.
		
		\State Perform a usual exploratory PCA to detect if a simple structure appears clearly in the loading matrix. 
		
		\State Obtain information on the variance explained by the principal components.
		
		\State Determine the number of disjoint components $Q$ according to the investigator's criteria based on the information obtained in Step 3.
		
		\State Calculate the $Q$ disjoint components according to the CBPSO DC method following
		
		\begin{enumerate}
			\item[5.1] Stablish the stopping conditions by considering the number of iterations to be performed.
			
			\item[5.2] Determine the number of particles, maximum and minimum inertia, as well as social and cognitive weights.
			
			\item[5.3] Apply Algorithm \ref{alg:2}.
			
		\end{enumerate}
		
		\State Repeat step 5 several times and choose the execution that presents the best fit.
		
		\State Analyze and report the obtained results.	
	\end{algorithmic}
\end{algorithm}

\subsection{Matrix generator with disjoint component structure}

A simulation algorithm is constructed to randomly generate a data matrix, with an ad hoc structure of easy interpretation, and then when applying a dimension reduction by disjoint main components, the CBPSO DC algorithm is able to detect it.

Let $x_1,\dots,x_p$ be original $p$ variables and $y_1, \dots,y_q$ be $q$ latent variables ($q<p$) with disjoint structure. Consider the linear combination given by
\begin{equation} \label{6.1)} 
\notag	
y_{i} =c_{1,i} x_{1} +\cdots +c_{p,i} x_{p}.
\end{equation} 
If the $m$ original consecutive  variables $x_j, x_{j+1},\dots,x_{j+(m-1)}$ are represented in the latent variable $y_i$, then the scalars $c_{j,i},c_{j+1,i},\dots, c_{j+(m-1),i}$ are defined as independent and uniform discrete random variables with support on the integers from 70 to 100, whereas the remaining scalars are defined similarly but with support on the integers from 1 to 30. This procedure is carried out for each $i$ from $1$ to $q.$ Keep in mind that each original variable must have a strong presence in a single latent variable.

\paragraph{Example 2} Suppose that $p=8$, $q=3$ and $\textrm{U}_{\textrm{d}}(a, b)$ is the discrete uniform distribution, with support on the integers from $a$ to $b$ ($a,b\in \mathbb{N}, \,a<b$). Note that he first linear combination is given by
$$
y_{1} =c_{1,1} x_{1} +c_{2,1} x_{2} +c_{3,1} x_{3} +c_{4,1} x_{4} +c_{5,1} x_{5} +c_{6,1} x_{6} +c_{7,1} x_{7} +c_{8,1} x_{8}.
$$
If $x_1,x_2,x_3,x_4$ are represented in $y_1$, then
$$
c_{1,1},c_{2,1},c_{3,1},c_{4,1} 
\sim  \textrm{U}_{\textrm{d}}(70,100), \quad
c_{5,1},c_{6,1},c_{7,1},c_{8,1} 
\sim  \textrm{U}_{\textrm{d}}(1,30),
$$
where ``iid'' denotes ``independent and identically distributed''. The second linear combination is defined as
$$
y_{2} =c_{1,2} x_{1} +c_{2,2} x_{2} +c_{3,2} x_{3} +c_{4,2} x_{4} +c_{5,2} x_{5} +c_{6,2} x_{6} +c_{7,2} x_{7} +c_{8,2} x_{8}.
$$
If $x_5,x_6,x_7$ are represented in $y_2$, then
$$
c_{5,2},c_{6,2},c_{7,2} 
\sim  \textrm{U}_{\textrm{d}}(70,100), \quad
c_{1,2},c_{2,2},c_{3,2},c_{4,2},c_{8,2} 
\sim  \textrm{U}_{\textrm{d}}(1,30),
$$
And the third linear combination is expressed as
$$
y_{3} =c_{1,3} x_{1} +c_{2,3} x_{2} +c_{3,3} x_{3} +c_{4,3} x_{4} +c_{5,3} x_{5} +c_{6,3} x_{6} +c_{7,3} x_{7} +c_{8,3} x_{8}.
$$
We only have left the original variable $x_8$. This must be represented in $y_3$. Then:
\begin{equation} \label{6.9)} 
\notag	
c_{8,3} \; \sim \; \; U_{d} \left(70,100\right) 
\end{equation} 
\begin{equation} \label{6.10)} 
\notag	
c_{1,3},c_{2,3},c_{3,3},c_{4,3},c_{5,3},c_{6,3},c_{7,3} \; \sim \; U_{d} \left(1,30\right)\; \; {\kern 1pt} IID 
\end{equation} 
In the mentioned example, the first four original variables have a strong presence in the first latent variable, the next three original variables do so in the second latent variable, and the last original variable has a strong representation in the third and last latent variable. We indicate this, in a general way, by the finite succession ${\left\{r_k\right\}}^q_{k=1}$, whose elements for the previous particular case are $r_1=4$, $r_2=3$ and $r_3=1$.The algorithm designed simulates a matrix with $n$ individuals and $p$ variables. The orthonormalization process of Gram-Schmidt is applied to this matrix to obtain the simulated data matrix $\bm{X}$. The algorithm that builds the random matrix $\bm{X}$ of size $n\times p$ should, in general terms, implement the mapping $\varphi $ defined as follows:
\begin{equation} \label{mapping}
\begin{array}{l} {\phi :\textrm{\textrm{N}}\times \textrm{\textrm{N}}\times \textrm{\textrm{N}}\times \textrm{\textrm{N}}^{q} \to M_{n\times p} } \\ {\quad \left(n,\; p,\; q,\left\{r_{k} \right\}_{k=1}^{q} \right)\mapsto {\it \phi }\left(n,\; p,\; q,\left\{r_{k} \right\}_{k=1}^{q} \right)} \end{array} 
\end{equation} 
subject to the constraints $n\ge p$ and $q<p$. In addition, it should be satisfied that:
\begin{equation} \label{6.12)} 
		\notag	
p=\mathop{\sum }\limits_{k=1}^{q} r_{k} 
\end{equation} 
The $\varphi $ mapping returns an $n\times p$ size matrix $\bm{X}$ that has the ad hoc structure mentioned above, which we hope is detected by the CBPSO CD algorithm.

\subsection{Simulation studies}

For the first simulation example this mapping was executed: 
\begin{equation} \label{mapping2}
		\notag	
\boldsymbol{\varphi }\left(100,\ 8,\ 3,\ \left\{4,\ 3,\ 1\right\}\right),
\end{equation}
that is, there are 100 individuals and 8 original variables. In addition, the latent structure is made up of three disjoint components: the first one has nonzero charges in the first four positions. The second disjoint component has nonzero charges at positions 5, 6 and 7. The last disjoint component has a nonzero charge at the last position. A data matrix $\bm{X}$ of size $100\times 8$ has been obtained. The CBPSO DC algorithm delivered the load matrix presented in Table \ref{sim11}.

\begin{table}[h]
	\centering
	\caption{Loading matrix $\bm{B}$ obtained by CBPSO DC and DPCA algorithms }
	\label{sim11}
	\resizebox{7cm}{!} { 
		\begin{tabular}{cccc p{0.5in}|p{0.76in}|p{0.76in}|p{0.76in}|} \toprule 
			~ & ${\boldsymbol{y}}_{\boldsymbol{1}}$\textbf{} & ${\boldsymbol{y}}_{\boldsymbol{2}}$\textbf{} & ${\boldsymbol{y}}_{\boldsymbol{3}}$\textbf{} \\ \hline 
			${\boldsymbol{x}}_{\boldsymbol{1}}$\textbf{} & 0.43413655 & 0 & 0 \\ 
			${\boldsymbol{x}}_{\boldsymbol{2}}$\textbf{} & 0.53903007 & 0 & 0 \\ 
			${\boldsymbol{x}}_{\boldsymbol{3}}$\textbf{} & 0.56700392 & 0 & 0 \\ 
			${\boldsymbol{x}}_{\boldsymbol{4}}$\textbf{} & 0.44663027 & 0 & 0 \\ 
			${\boldsymbol{x}}_{\boldsymbol{5}}$\textbf{} & 0 & 0.57919604 & 0 \\ 
			${\boldsymbol{x}}_{\boldsymbol{6}}$\textbf{} & 0 & 0.56750625 & 0 \\ 
			${\boldsymbol{x}}_{\boldsymbol{7}}$\textbf{} & 0 & 0.58520817 & 0 \\ 
			${\boldsymbol{x}}_{\boldsymbol{8}}$\textbf{} & 0 & 0 & 1 \\ \midrule 
			\textbf{\% VAR} & 35.54\% & 33.01\% & 29.26\% \\ \bottomrule
	\end{tabular} }
\end{table}

The CBPSO DC algorithm was executed 100 times with 50 particles and 10 iterations as the stop criterion. In all 100 executions the same solution was reached, shown in table \ref{sim11}, needing two or three iterations at the most. The CBPSO DC algorithm was able to detect the ad hoc structure contained in the data. The fit obtained when applying disjoint principal components is 0.021859878 and the total variance explained by the model is 97.81\%. The DPCA algorithm was also executed 100 times and the same solution given in table \ref{sim11} was reached 100\% of the time. A usual PCA was also performed on the same data matrix and the load matrix given in Table \ref{sim12} was obtained. Note that the usual PCA does not clearly detect the underlying structure in the data. For example, what happens with the second original variable $x_2$ is noteworthy: it cannot be concluded with certainty in which component it has a strong presence. The same can be said of the last original variable $x_8$.

\begin{table}[h]
	\centering
	\caption{Loading matrix $\bm{B}$ obtained by usual PCA algorithm}
	\label{sim12}
	\resizebox{7cm}{!} { 
		\begin{tabular}{cccc p{0.7in}|p{1in}|p{1in}|p{1in}|} \toprule 
			~ & ${\boldsymbol{y}}_{\boldsymbol{1}}$\textbf{} & ${\boldsymbol{y}}_{\boldsymbol{2}}$\textbf{} & ${\boldsymbol{y}}_{\boldsymbol{3}}$\textbf{} \\ \midrule 
			${\boldsymbol{x}}_{\boldsymbol{1}}$\textbf{} & 0.39512468 & -0.09020236 & 0.14824807 \\ 
			${\boldsymbol{x}}_{\boldsymbol{2}}$\textbf{} & 0.39775809 & -0.00322755 & 0.40788776 \\ 
			${\boldsymbol{x}}_{\boldsymbol{3}}$\textbf{} & 0.44254179 & -0.17190775 & 0.31755350 \\ 
			${\boldsymbol{x}}_{\boldsymbol{4}}$\textbf{} & 0.35239424 & -0.11448361 & 0.25061667 \\ 
			${\boldsymbol{x}}_{\boldsymbol{5}}$\textbf{} & 0.26682407 & 0.46011771 & -0.25159988 \\ 
			${\boldsymbol{x}}_{\boldsymbol{6}}$\textbf{} & 0.30662152 & 0.39441186 & -0.26207760 \\ 
			${\boldsymbol{x}}_{\boldsymbol{7}}$\textbf{} & 0.35705213 & 0.28282245 & -0.38998176 \\ 
			${\boldsymbol{x}}_{\boldsymbol{8}}$\textbf{} & -0.27007773 & 0.70847497 & 0.60326462 \\ \midrule 
			\textbf{\% VAR} & 39.81\% & 32.27\% & 27.92\% \\ \bottomrule
	\end{tabular} }
\end{table}

A second simulation study is provided below as an example. This example was designed to compare the performance of the CBPSO DC algorithms and the DPCA algorithm when the matrix size is larger. The simulation algorithm implements the mapping \eqref{mapping} given by
\begin{equation} \label{mapping3}
		\notag	
\boldsymbol{\varphi }(200, 200, 3, \{50,70, 80\})
\end{equation}
and an $\bm{X}$ data matrix of size $200\times 200$ and three disjoint components was obtained. After executing both algorithms 100 times each, the results obtained are summarized in Table \ref{sim13}.

\begin{table}[h]
	\centering
	\caption{Results summary of the simulation study II}
	\label{sim13}	
	\resizebox{7cm}{!} { 
		\begin{tabular}{lcc p{3.3in}|p{0.74in}|p{0.73in}|} \toprule 
			~ & \textbf{CBPSO DC} & \textbf{DPCA} \\ \midrule 
			\textbf{AVERAGE EXECUTION TIME (MIN)} & 4.6 & 10.2 \\ 
			\textbf{BETTER EXECUTION TIME (MIN)} & 4.2 & 9.8 \\
			\textbf{WORST EXECUTION TIME (MIN)} & 4.8 & 10.4 \\ \midrule 
			\textbf{SUCCESS RATE} & 100\% & 93\% \\ \bottomrule 
	\end{tabular}}
\end{table}

Both algorithms found the best solution, with a fit of 0.02543703 and a total explained variance of 97.46\%, with different success rates. The CBPSO DC algorithm was executed with the same input parameters as in the previous computational experiments, and the best time reached was 4.2 minutes in 6 iterations. On the other hand, for the DPCA algorithm the best time reached was 9.8 minutes in two iterations. The last row of the table tells us that the DPCA was trapped in a local optimum in 7 of the 100 executions. It is important to highlight that the DPCA algorithm performs an exhaustive search, since starting from the binary matrix $\bm{{V}}$ that it initially generates randomly, it begins to move by number and by column the number one in all possible positions. The greater the number of original variables and latent variables, the longer processing time is required by this algorithm. The CBPSO DC algorithm is better suited for large arrays, since it has the flexibility to allow an adjustment in its parameters that allow a better search strategy. The DPCA algorithm has only one parameter, tolerance, which is used as a stop criterion. With respect to the percentages of success, the DPCA algorithm has a greater probability of being trapped in a local optimum, since it begins its processing with a single binary matrix $\bm{{V}}$, which, as said, is modified by displacements of the one row and column, not necessarily leading to an optimal matrix. As mentioned in \cite{Macedo2015}, the algorithm can be trapped in different local optima and it is recommended to execute it several times. On the other hand, the CBPSO DC algorithm starts with a matrix $\bm{{V}}$ for each particle, and it is the joint and intelligent search of the particles which allows the CBPSO DC to have a lower probability of being trapped in a local optimum.

\subsection{Convergence analysis}

\begin{figure}[h]
	\centering
	\caption{Convergence analysis for SIMULATION STUDY II}
	\label{fig:sim21}
	\includegraphics[width=0.5\textwidth]{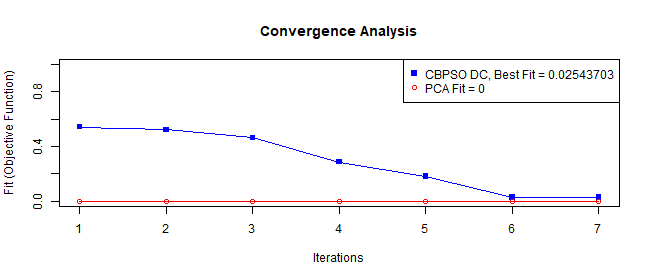}
\end{figure}

To analyze the convergence of the proposed algorithm, from a comparative point of view, we contrast it with the usual PCA. If the mathematical optimization model for obtaining disjoint components (derivative model) is relaxed, eliminating precisely the constraint for the calculation of disjoint components, the minimization model for the calculation of classical principal components (primary model) is obtained. Therefore, the fit of a classic PCA is always worse than the fit of a PCA with disjoint components. Figure \ref{fig:sim21} shows a blue curve that relates the number of iterations of the CBPSO DC algorithm with the fit (value of the objective function). The red line, parallel to the horizontal axis, shows the fit of a classic PCA. For the second simulation study the fit of a classic PCA is so small that the R software represents that fit with a zero. It is observed that the best fit obtained by the CBPSO DC algorithm in one of the executions is 0.02543703 in 6 iterations of a maximum of 7 iterations.

\section{Empirical illustrations} \label{sec:6}

\subsection{The case of trait meta-mood scale 24}

Recent studies show the importance of emotional intelligence in the use of learning strategies. A questionnaire that measures this construct is called TMMS24 as detailed in \cite{Vega-Hernandez2017}. This questionnaire consists of a set of 24 questions to assess emotional states, each of them measured on a Likert scale of 1 (No agreement) to 5 points (Strongly agree). This instrument follows from the Trait-Meta Mood Scale (TMMS) of the Salovey and Mayer research group \cite{Sanchez-Garcia2016}. This questionnaire has been used on purpose, as it is a validated questionnaire, in which its latent dimensions are clearly distinguished. TMMS 24 contains three dimensions or factors: \texttt{Emotional attention} (questions from P1 to P8); \texttt{Emotional clarity} (questions from P9 to P16); \texttt{Emotional repair} (questions from P17 to P24). In this computational experiment, the TMMS24 questionnaire was administered to 249 students of the University of Salamanca, Spain, and the data was processed first with the usual PCA and then with the two disjoint methods treated, DPCA and CBPSO DC. Table \ref{sim21} shows the loads provided by the usual PCA. As the values of the factor loads are given, it turns out to be a very complicated task to interpret each of the latent variables of the TMMS24 questionnaire in terms of the pre-established dimensions, because there are no remarkably high values in absolute value that differentiate them from the rest.

\begin{table}[h]
	\centering
	\caption{Usual PCA for TMMS24 questionnaire}
	\label{sim21}
	\resizebox{14cm}{!} {
		\begin{tabular}{cccccccccccc p{0.21in}|p{0.5in}|p{0.5in}|p{0.5in}|p{0.26in}|p{0.5in}|p{0.5in}|p{0.5in}|p{0.26in}|p{0.5in}|p{0.5in}|p{0.5in}|} \toprule
			~ & \textbf{PCA1} & \textbf{PCA2} & \textbf{PCA3} & ~ & \textbf{PCA1} & \textbf{PCA2} & \textbf{PCA3} & ~ & \textbf{PCA1} & \textbf{PCA2} & \textbf{PCA3} \\ \midrule 
			P1 & 0.23254 & -0.23237 & 0.08136 & P9 & 0.22998 & 0.08135 & -0.36558 & P17 & 0.16413 & 0.33062 & 0.21872 \\ 
			P2 & 0.22928 & -0.25397 & 0.09367 & P10 & 0.25446 & 0.04687 & -0.33838 & P18 & 0.14701 & 0.34381 & 0.24822 \\
			P3 & 0.23245 & -0.26459 & 0.15229 & P11 & 0.23467 & 0.08685 & -0.26817 & P19 & 0.15706 & 0.30838 & 0.27075 \\ 
			P4 & 0.27546 & -0.19912 & 0.04398 & P12 & 0.18133 & 0.00459 & -0.21184 & P20 & 0.16961 & 0.33375 & 0.22150 \\ 
			P5 & 0.14383 & -0.26921 & 0.13748 & P13 & 0.20757 & 0.00405 & -0.14915 & P21 & 0.16925 & 0.17367 & 0.10622 \\
			P6 & 0.18181 & -0.23168 & 0.17093 & P14 & 0.22040 & 0.09600 & -0.30302 & P22 & 0.19467 & 0.09546 & 0.11455 \\ 
			P7 & 0.18510 & -0.23702 & 0.22662 & P15 & 0.20857 & 0.04957 & -0.16494 & P23 & 0.10024 & 0.08436 & 0.07860 \\ 
			P8 & 0.25175 & -0.20199 & 0.16564 & P16 & 0.21579 & 0.05754 & -0.21777 & P24 & 0.21486 & 0.19223 & 0.12857 \\ \bottomrule
		\end{tabular}
	}
\end{table}

When using the DPCA and CBPSO DC methods, the best solution found proved to be exactly the same and is presented in Table \ref{sim22}.

\begin{table}[h]
	\centering
	\caption{Best solution}
	\label{sim22}
	\scriptsize
	\resizebox{14cm}{!} {
		\begin{tabular}{cccccccccccc p{0.16in}|p{0.39in}|p{0.319in}|p{0.319in}|p{0.215in}|p{0.39in}|p{0.39in}|p{0.39in}|p{0.215in}|p{0.319in}|p{0.59in}|p{0.39in}|} \toprule
			\textbf{~} & \textbf{COMP1} & \textbf{COMP2} & \textbf{COMP3} & \textbf{~} & \textbf{COMP1} & \textbf{COMP2} & \textbf{COMP3} & \textbf{~} & \textbf{COMP1} & \textbf{COMP2} & \textbf{COMP3} \\\midrule
			P1 & 0.35227 & 0 & 0 & P9 & 0 & 0 & -0.41062 & P17 & 0 & 0.42247 & 0 \\ 
			P2 & 0.36616 & 0 & 0 & P10 & 0 & 0 & -0.42258 & P18 & 0 & 0.42936 & 0 \\ 
			P3 & 0.38711 & 0 & 0 & P11 & 0 & 0 & -0.38266 & P19 & 0 & 0.41784 & 0 \\ 
			P4 & 0.36503 & 0 & 0 & P12 & 0 & 0 & -0.28486 & P20 & 0 & 0.42935 & 0 \\ 
			P5 & 0.31388 & 0 & 0 & P13 & 0 & 0 & -0.28216 & P21 & 0 & 0.28538 & 0 \\ 
			P6 & 0.32747 & 0 & 0 & P14 & 0 & 0 & -0.38453 & P22 & 0 & 0.24283 & 0 \\ 
			P7 & 0.34493 & 0 & 0 & P15 & 0 & 0 & -0.30018 & P23 & 0 & 0.15940 & 0 \\ 
			P8 & 0.36605 & 0 & 0 & P16 & 0 & 0 & -0.32812 & P24 & 0 & 0.33528 & 0 \\ \bottomrule
		\end{tabular}
	}	
\end{table}

It can be seen that the disjoint algorithms can classify the 24 questions of the TMMS24 questionnaire into three clearly defined groups, which correspond precisely to the dimensions of the questionnaire. The disjoint algorithms distinguish precisely the dimensions that the questionnaire claims to have, without creating spurious and non-existent dimensions. In addition, in the two disjoint methods the fit, or relative error of the approximation, for this solution is equal to 0.4450664, while for the usual PCA it is 0.4306895. There is a slight loss of fit, but in return there is a gain in interpretation. The proportion or percentage of variance explained by each disjoint component is presented in Table \ref{sim23} and turns out to be the same for both methods compared. These percentages are calculated by means of \eqref{ZEqnNum234730}.

\begin{table}[h]
	\centering
	\caption{The proportion of variance explained by disjoint components}
	\label{sim23}
	\resizebox{7cm}{!} { 
		\begin{tabular}{ccccc p{0.65in}|p{0.643in}|p{0.643in}|p{0.643in}|p{0.643in}|} \toprule 
			~ & \textbf{COMP1} & \textbf{COMP2} & \textbf{COMP3} & \textbf{TOTAL} \\ \midrule 
			\textbf{\% VAR} & 20.45\% & 18.23\% & 16.81\% & \textbf{55.49\%} \\ \bottomrule
	\end{tabular} }
\end{table}

The proportion of variance explained by the three disjoint components considered is approximately 1.5\% less than the variance explained by the first three principal components calculated in the usual way, as can be seen from Table \ref{sim24}.

\begin{table}[h]
	\centering
	\caption{ The proportion of variance explained by usual PCA}
	\label{sim24}	
	\resizebox{7cm}{!} { 
		\begin{tabular}{ccccc p{0.62in}|p{0.63in}|p{0.63in}|p{0.63in}|p{0.63in}|} \toprule 
			~ & \textbf{COMP1} & \textbf{COMP2} & \textbf{COMP3} & \textbf{TOTAL} \\ \midrule 
			\textbf{\% VAR} & 27.05\% & 19.27\% & 10.61\% & \textbf{56.94\%} \\ \bottomrule 
	\end{tabular} }
\end{table}

Where the improvement of the CBPSO DC algorithm with respect to the DPCA algorithm can be seen is in the proportion of successes in reaching the best solution. The CBPSO DC algorithm explores the set of feasible solutions more thoroughly, which explains why it is not trapped in a local optimum. One hundred executions were carried out for each of the exposited methods. The DPCA method reached the best solution (shown in Table \ref{signif}) in 98\% of the cases, while in the CBPSO DC method it did so in 100\% of the cases.

\subsection{The case of general self-eficacy scale}

The concept of general self-efficacy refers to the way in which an individual perceives their ability to perform new or difficult tasks, and to cope with difficulties. This construct is quantified through the General Self Efficacy Scale (GSE), see \cite{Scholz2002}, that establishes that the latent structure is one-dimensional and universal. That is, the 10 items that make up the scale form a factor that can be used in a generalized way in any country. However, as noted in \cite{Villegas2018} this scale is not one-dimensional or universal, a conclusion that is confirmed by the analysis by disjoint components presented below. 

The GSE scale consists of 10 items evaluated with a four-point Likert scale, according to the following categorization: 1 Not quite true, 2 Almost true, 3 Moderately true and 4 Exactly true. The meaning of each item is detailed in Table \ref{signif}.

\begin{table}[h]
	\centering
	\caption{Meaning of the variables in GSE Scale}
	\label{signif}
	\resizebox{11cm}{!} {
		\begin{tabular}{ccccll}
			\toprule
			\textbf{Item}&\textbf{Meaning} \\ \midrule 
			Self1 & I can always handle to solve difficult problems if I try hard enough. \\ 
			Self2 & If someone opposes me, I can find means and ways to get what I want. \\ 
			Self3 & It is easy for me to stick to my aims and accomplish my goals. \\ 
			Self4 & I am confident that I could deal efficiently with unexpected events. \\ 
			Self5 & Thanks to my resourcefulness, I know how to handle unforeseen situations. \\ 
			Self6 & I can solve most problems if I invest the necessary effort. \\ 
			Self7 & I can remain calm when facing difficulties because I can rely on my coping abilities. \\ 
			Self8 & When I am confronted with a problem, I can usually find several solutions. \\ 
			Sel9 & If I am in trouble, I can usually think of something to do. \\ 
			Self10 & No matter what comes my way, I am usually able to handle it. \\ \bottomrule
		\end{tabular}
	}
\end{table}

For example, a database was used, with the 10 items and 19,719 individuals, which is available at: http://userpage.fu-berlin.de/$\mathrm{\sim}$health/selfscal.htm.

The CBPSO DC algorithm shows three disjoint components with a fit of 0.02820508 and a total explained variance of 58.93\%. This result is similar to the one obtained by \cite{Villegas2018}. It can be seen in table \ref{sim31} that the variables Self1 and Self6 are found in the second component. The Self2 variable is the only one found in the third component. The other variables are found in the first component. The information provided by these items is not collected in a single latent variable, contrary to what the authors of the scale affirm. It is important to conclude then that the GSE scale is not one-dimensional: it cannot be described through a single factor. See table \ref{sim31}.

\begin{table}[h]
	\centering
	\caption{GSE Scale principal components by CBPSO DC and DPCA}
	\label{sim31}
	\resizebox{6cm}{!} {
		\begin{tabular}{cccc p{0.5in}|p{0.6in}|p{0.6in}|p{0.6in}|} \toprule
			~ & \textbf{COMP1} & \textbf{COMP2} & \textbf{COMP3} \\ \midrule 
			\textbf{Self1} & 0 & 0.71776986 & 0 \\ 
			\textbf{Self2} & 0 & 0 & 1 \\ 
			\textbf{Self3} & 0.36286421 & 0 & 0 \\ 
			\textbf{Self4} & 0.36971158 & 0 & 0 \\ 
			\textbf{Self5} & 0.37589985 & 0 & 0 \\ 
			\textbf{Self6} & 0 & 0.69628042 & 0 \\ 
			\textbf{Self7} & 0.38156436 & 0 & 0 \\ 
			\textbf{Self8} & 0.38424172 & 0 & 0 \\ 
			\textbf{Self9} & 0.3867569 & 0 & 0 \\ 
			\textbf{Self10} & 0.38409408 & 0 & 0 \\ \midrule 
			\textbf{\% EV} & 35.4\% & 12.8\% & 10.73\% \\ \bottomrule 
	\end{tabular}}
\end{table}

This large data matrix is also analyzed with the DPCA algorithm. After performing 100 executions with both algorithms, the results obtained indicate that both algorithms found the best solution, presented in Table \ref{sim31}, with different success rates. The CBPSO DC algorithm was executed with 500 particles, 30 iterations maximum, 1.5 social weight and cognitive weight, and an inertia that varies linearly from 0.5 to 3. The best time reached by the CBPSO DC algorithm was 1.55 minutes in 4 iterations. In addition, for the DPCA algorithm the best time reached was 0.07 minutes in one iteration. However, and what we consider of vital importance, is that of the 100 executions, DPCA was trapped in a local optimum in 43\% of the executions, while CBPSO reached the best solution 100\% of the time.

\subsection{Convergence analysis}

\begin{figure}
	\centering
	\caption{Convergence Analysis TMMS24 Test}
	\label{fig:tmms241}
	\includegraphics[width=0.6\textwidth]{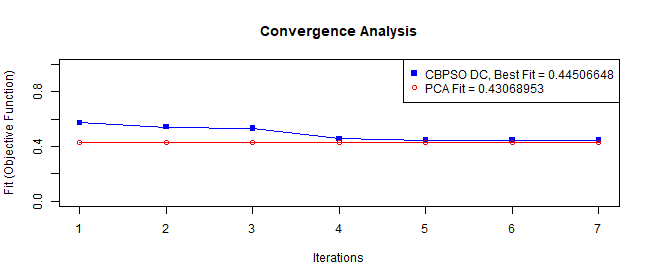}
\end{figure}

As in section 5.4, we compare the convergence of the CBPSO DC algorithm with the usual PCA. Figure \ref{fig:tmms241} shows one of the many executions of the CBPSO DC algorithm with the TMMS24 questionnaire data. Although 7 iterations are shown, in the fifth, the lowest fit equal to 0.4450664 was obtained. The red horizontal line shows the fit of a classic PCA whose value is 0.4306895, which represents a lower bound for the blue curve.

\section{Conclusions and recommendations}\label{sec:7}

In this article, we have proposed a new algorithm for calculating the principal components, which replaces the current implementations with a binary optimization with constrains using PSO. Because the components are of norm one and disjoint, the matrix of factorial loads is orthogonal, which can be interpreted as a rotation of the space of individuals. This allows preserving the original inertia in the rotated space, and thus using, in a natural way, the variance of the new variables as an indicator of the variance explained by the disjoint components. By replacing the local search used in the DPCA algorithm with the intelligent search of the CBPSO DC algorithm, the set of feasible solutions is better explored, allowing high-quality solutions to be reached regardless of the size of the matrix. On the other hand, the particles have shared memory which allows them to ignore routes that lead to solutions that represent local optima, something which is seen in the percentage of successes in obtaining the best solution in the empirical illustration II.
The DPCA method seeks the global optimum through successive local optimal choices, that is, it is an algorithm of the type known as greedy algorithms. For small instances a greedy algorithm usually delivers the optimal solution, but in larger instances, it is usually trapped in a local optimum. In general terms, greedy algorithms simply provide a good solution. The quality of this solution usually degrades as the size of the data set increases. This is shown in the two empirical illustrations in section 6. In the example of the TMMS24 questionnaire, 24 variables and 249 individuals, the percentage of correct answers to the best solution is 98\%. While in the example of the GSE questionnaire, the size of the data set increases considerably, 10 variables and 19719 individuals, the percentage of times that the DPCA algorithm reaches the best solution also decreases considerably to 57\%, compared to the 100\% effectiveness of the CBPSO DC algorithm in both cases. It should be mentioned that the fit obtained with the usual PCA is slightly better than the fit obtained with the disjoint principal components. However, what is lost in fit is compensated for by obtaining a factor structure that is easier to interpret. 

\section{Acknowledgements}\label{sec:8}

This work has been supported by ESPOL Polytechnic University, Escuela Superior Polit\'ecnica del Litoral, ESPOL. Guayaquil, Ecuador.



 
\bibliographystyle{elsarticle-harv}
\bibliography{biblio}

\end{document}